\documentclass[preprint,12pt]{elsarticle}

\usepackage{amsmath,amsfonts,amssymb}
\usepackage{graphicx}
\usepackage{setspace}
\usepackage{tocloft}
\usepackage{textcomp}
\usepackage{color,soul}
\usepackage{subcaption}
\usepackage{float}
\usepackage[skip=8pt]{caption}
\usepackage[colorlinks=true, allcolors=blue]{hyperref}
\usepackage[table,xcdraw,dvipsnames]{xcolor}
\usepackage{lineno}
\usepackage{array}
\usepackage{listings}
\usepackage{mathptmx}
\usepackage{epsfig}
\usepackage{rotating}
\usepackage{makeidx}
\usepackage{url}
\usepackage{booktabs}
\usepackage{tabularx}
\usepackage{blindtext}
\usepackage{times}
\usepackage{nccmath}
\usepackage{mwe}
\usepackage{acro}
\usepackage{adjustbox}
\usepackage{colortbl}
\usepackage{relsize}
\usepackage{pifont}
\usepackage{multirow}
\usepackage{multicol}
\usepackage{nicematrix}
\usepackage{bm}
\usepackage{makecell}
\usepackage{tabu}
\usepackage[capitalize]{cleveref}
\DeclareMathSymbol{\shortminus}{\mathbin}{AMSa}{"39}

\usepackage{color}
\definecolor{dkgreen}{rgb}{0,0.6,0}
\definecolor{gray}{rgb}{0.5,0.5,0.5}
\definecolor{mauve}{rgb}{0.58,0,0.82}

\colorlet{Mycolor1}{green!10!orange}


\usepackage{lineno}

\journal{Meta-Radiology}

\begin{document}

\begin{frontmatter}

\title{Fast-RF-Shimming: Accelerate RF Shimming in 7T MRI using Deep Learning}

\author[a]{Zhengyi Lu}
\author[b,c]{Hao Liang}
\author[b,c]{Ming Lu}
\author[d]{Xiao Wang}
\author[b,c,a]{Xinqiang Yan}
\author[e,a]{Yuankai Huo}

\affiliation[a]{organization={Department of Electrical and Computer Engineering, Vanderbilt University},
            city={Nashville},
            state={TN},
            country={USA}}

\affiliation[b]{organization={Vanderbilt University Institute of Imaging Science, Vanderbilt University Medical Center},
            city={Nashville},
            state={TN},
            country={USA}}

\affiliation[c]{organization={Department of Radiology and Radiological Sciences, Vanderbilt University Medical Center},
            city={Nashville},
            state={TN},
            country={USA}}

\affiliation[d]{organization={Computational Science and Engineering Division, Oak Ridge National Laboratory},
            city={Oak Ridge},
            state={TN},
            country={USA}}

\affiliation[e]{organization={Department of Computer Science, Vanderbilt University},
            city={Nashville},
            state={TN},
            country={USA}}

\begin{abstract}
Ultrahigh field (UHF) Magnetic Resonance Imaging (MRI) offers an elevated signal-to-noise ratio (SNR), enabling exceptionally high spatial resolution that benefits both clinical diagnostics and advanced research. However, the jump to higher fields introduces complications, particularly transmit radiofrequency (RF) field ($B_{1}^{+}$) inhomogeneities, manifesting as uneven flip angles and image intensity irregularities. These artifacts can degrade image quality and impede broader clinical adoption. Traditional RF shimming methods, such as Magnitude Least Squares (MLS) optimization, effectively mitigate $B_{1}^{+}$ inhomogeneity, but remain time-consuming. Recent machine learning approaches, including RF Shim Prediction by Iteratively Projected Ridge Regression and other deep learning architectures, suggest alternative pathways. Although these approaches show promise, challenges such as extensive training periods, limited network complexity, and practical data requirements persist.  In this paper, we introduce a holistic learning-based framework called Fast-RF-Shimming, which achieves a 5000× speed-up compared to the traditional MLS method. In the initial phase, we employ random-initialized Adaptive Moment Estimation (Adam) to derive the desired reference shimming weights from multi-channel $B_{1}^{+}$ fields. Next, we train a Residual Network (ResNet) to map $B_{1}^{+}$ fields directly to the ultimate RF shimming outputs, incorporating the confidence parameter into its loss function. Finally, we design Non-uniformity Field Detector (NFD), an optional post-processing step, to ensure the extreme non-uniform outcomes are identified. Comparative evaluations with standard MLS optimization underscore notable gains in both processing speed and predictive accuracy, which indicates that our technique shows a promising solution for addressing persistent inhomogeneity challenges.
\end{abstract}




\end{frontmatter}

\section{Introduction}
In Magnetic Resonance Imaging (MRI), RF shimming or Parallel Transmission (PTx) has emerged as a critical technology for producing high-quality images at ultra high fields (UHF)~\cite{Zhu2004_add, Katscher2006_add}. As the static field strength rises, the RF wavelength becomes comparable to tissue dimensions, causing destructive interferences~\cite{VandeMoortele2005_add}. These interferences induce inhomogeneous $B_{1}^{+}$ fields, leading to inconsistent flip angles and abnormal image intensities. If not addressed, such non-uniformity reduces overall image quality at UHF and hinders broader adoption in medical practice~\cite{rv_snr}.

Traditional methods like Magnitude Least Squares (MLS) optimization have been widely used for RF shimming, improving field uniformity but requiring time-intensive computations and subject-specific calibrations~\cite{4Setsompop2008, 10Padormo2016, 11Kilic2024}.

Recent advancements in machine learning have shown promise in addressing these issues, enabling faster predictions of RF shimming weights~\cite{12Plumley2022, 20Mirfin2018, 21Shin2021}. However, many of these approaches struggle with computational efficiency, artifact detection, and architectural limitations, leaving significant gaps in practical applications. 

\begin{figure*}[h]
\centering
\includegraphics[width=0.85\linewidth]{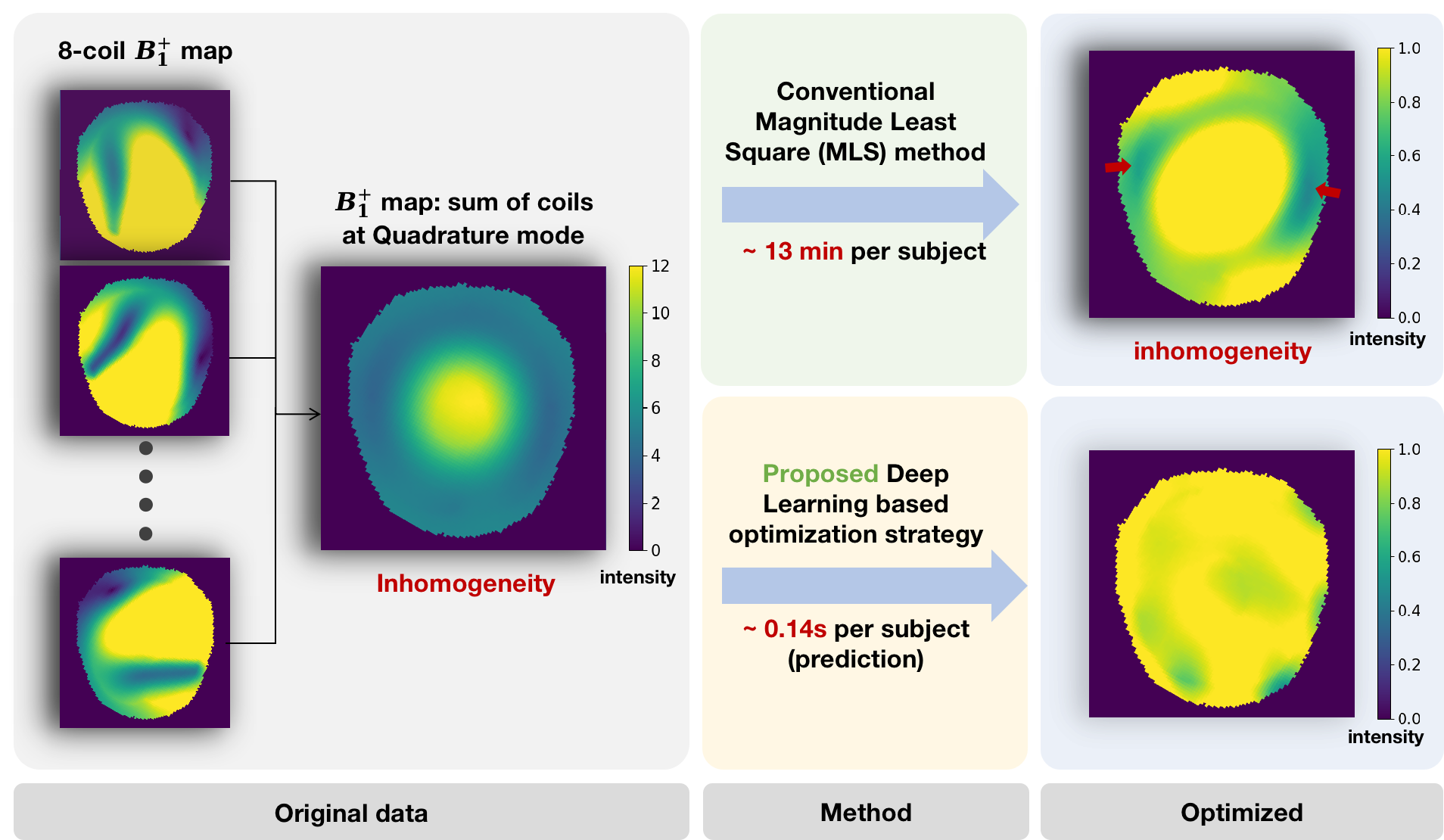}
\caption{The $B_{1}^{+}$ field inhomogeneity problem addressed in RF shimming design and main advantages of our proposed strategy over conventional MLS method. The block to the left presents the 8-coil $B_{1}^{+}$ data we started with. The two central blocks show our method and conventional MLS method with measured runtime over one subject based on MPRAGE~\cite{Mugler1990} which is estimated from testing prediction. On the right are two optimization results on the same slice picked from testing cases, which shows a contrast of uniformity over the $B_{1}^{+}$ field between the two methods.}
\label{fig:Figure_1_clean}
\end{figure*}

In this paper, we introduce a holistic learning-based framework called Fast-RF-Shimming, as shown in Fig.~\ref{fig:Figure_1_clean}, which achieves a 5000× speed-up in RF shimming compared to traditional MLS optimization methods. To achieve this, we develop a random-initialized Adaptive Moment Estimation (Adam) algorithm~\cite{17Kingma2014} to determine the desired reference RF weights as training upper bounds from multichannel $B_{1}^{+}$ fields across slices. Next, we adopt a Residual Network (ResNet) framework introduced by He \textit{et al.}~\cite{13He2016}, which learns to map the $B_{1}^{+}$ field directly to the RF shimming upper bound outputs. Finally, we design Non-uniformity Field Detector (NFD), an optional post-processing step, to ensure the extreme non-uniform outcomes are identified. We compare our method with traditional MLS optimization and Unsupervised CNN (uCNN)~\cite{11Kilic2024} in terms of Root Mean Square Error (RMSE) and computational time, evaluating performance slice-by-slice. The contribution of this paper is threefold. 

\vspace{-1em}

\begin{itemize} 
\item We introduce a holistic learning-based framework called Fast-RF-Shimming, which achieves a 5000× speed-up in RF shimming compared to traditional MLS optimization methods.
\item We develop a random-initialized Adam to derive high-quality training data (regarded as upper-bound) as reference RF shimming weights for training the learning based framework, mitigating the risk of local minima commonly seen in MLS~\cite{17Kingma2014}.
\item A ResNet architecture~\cite{13He2016} is leveraged to efficiently learn residual functions from $B_{1}^{+}$ magnetic fields to complex shimming weights for each transmit channel. 
\end{itemize}

\section{Related work}
\subsection{Traditional RF Shimming Techniques}
One widely used approach to \(B_{1}^{+}\) field uniformity is RF shimming with an n-element coil array, applied either on specific slices or across the entire imaging volume~\cite{Mao2006}. In magnitude-only scenarios, measuring each coil’s \(B_{1}^{+}\) map for each subject allows for phase and amplitude calibration specific to the scan setup~\cite{4Setsompop2008}. Setsompop et al.\ introduced a Magnitude Least Squares (MLS) optimization method for parallel RF excitation at 7 Tesla, leveraging an eight-channel transmit array to improve the \(B_{1}^{+}\) magnitude profile. This approach parallels phase retrieval techniques widely applied in other research fields~\cite{4Setsompop2008, 5Guerin2014, 6Kerr2007}. 

Further refinements to MLS optimization have been made in subsequent studies to enhance magnetization uniformity~\cite{10Padormo2016, 9Grissom2012}. However, these methods often require patient presence during scanner-based computations, significantly extending the time needed for deriving RF shimming weights~\cite{11Kilic2024, 8Cao2016}. This limitation highlights the need for more efficient approaches that eliminate the requirement for patient presence during optimization.

\subsection{Machine Learning Approaches}
Deep learning has been widely adopted in the medical imaging field, demonstrating strong capabilities in image reconstruction and segmentation~\cite{rv_zhu2023, rv_qu2025}.Recently, machine learning-based methods have been proposed to address \(B_{1}^{+}\) inhomogeneity by learning the mapping from the scanned field to the desired magnetization~\cite{20Mirfin2018, 21Shin2021, Ianni2018}. A notable example is the RF Shim Prediction by Iteratively Projected Ridge Regression (PIPRR) introduced by Ianni et al., which combines training shim designs with interpolation across learned shims~\cite{Ianni2018}. Although effective for slice-by-slice predictions, PIPRR has limitations, such as long training times, often requiring five days without parallelization across slices.

Another machine learning-based effort employs deep learning to predict \(B_{1}^{+}\) distributions after within-slice motion, allowing real-time pulse adjustments and reducing motion-related excitation errors~\cite{12Plumley2022}. Furthermore, an unsupervised convolutional neural network (CNN) approach guided by a physics-driven loss function has been used to minimize discrepancies between Bloch simulation results and target magnetization~\cite{11Kilic2024}.

\section{Method}
\subsection{Magnitude Least Squares Optimization}
MLS optimization specifically aims to minimize the discrepancy between the desired and actual magnitudes of the $B_{1}^{+}$ field within the imaging volume~\cite{4Setsompop2008}. In contrast to methods focusing on phase or real/imaginary components~\cite{15LeRoux1998, 16Collins2005}, MLS targets the field’s magnitude alone. The $B_{1}^{+}$ contribution of each coil element is typically modeled using electromagnetic simulations, and the aggregate $B_{1}^{+}$ field at any location in the imaging volume is the vector sum of the fields from all coils. After modeling, the optimization problem is formulated as~\cite{4Setsompop2008, 14Zheng2012}:

\begin{equation}\label{b}
\begin{split}
    b(t) = \arg\min_b \left\{ \left\| \left| Ab \right| - m \right\|_w^2 + \lambda \left\| b \right\|^2 \right\}
\end{split}
\end{equation}

\noindent where $A$ is the matrix of $B_{1}^{+}$ field values for each coil at every spatial location, $b$ denotes the coil weight vector to be determined, $m$ represents the desired magnetic field map, $w$ is a mask defining the region of interest, and $\lambda$ is the regularization term balancing RF power and excitation errors~\cite{14Zheng2012}.

\begin{figure*}[t]
\centering
\includegraphics[width=1\linewidth]{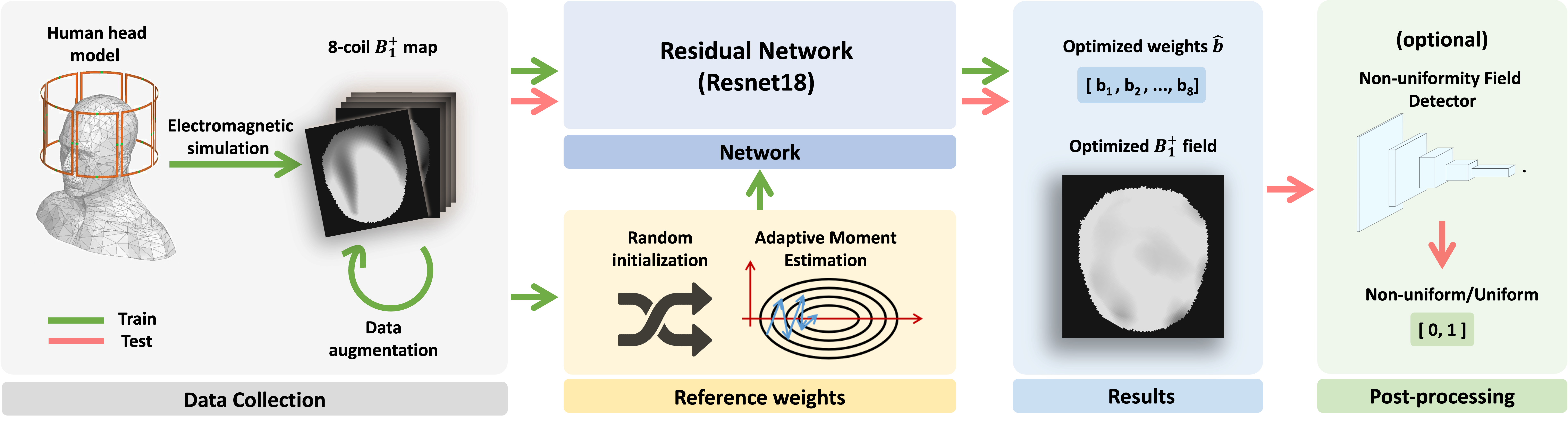}
\caption{The procedures of training and prediction of our proposed optimization strategy. The strategy starts with simulations to get the desired $B_{1}^{+}$ as inputs of the Residual Network (ResNet18). Data augmentation is then applied. Adaptive Moment Estimation~\cite{17Kingma2014} is used to calculate the reference weights of coils as targets in training. Next, predictions are made by applying the testing data into the trained DL model. Finally, an optional post-processing step with the NFD is designed to classify non-uniform and uniform output fields.}
\label{fig:Figure_2_clean}
\end{figure*}

\subsection{Quadrature Mode and Random Initialization}
Both the random-initialized Adaptive Moment Estimation and the MLS compression approach begin with coil elements arranged in a quadrature configuration, following the simulation setup illustrated in Fig.~\ref{fig:Figure_2_clean}. The eight RF coils, each angled at 45 degrees around the subject~\cite{19Ibrahim2005}, serve as the baseline arrangement for optimization.

Since our data are derived from practical EM simulation results, it is not feasible to obtain precise ground-truth values for the RF shimming weights. Therefore, best-performing weights from Adaptive Moment Estimation algorithm~\cite{17Kingma2014} are chosen and used as near-truth upper bounds for training. The performance and convergence of Adam are strongly influenced by random initialization, as emphasized by Kingma and Ba, who noted its impact on early optimization dynamics~\cite{17Kingma2014}, and Reddi et al., who highlighted that improper initialization can lead to poor generalization or divergence~\cite{Reddi2018}. To enhance the likelihood of finding an optimal solution, we employ random initialization within Adaptive Moment Estimation when deriving the reference RF shimming weights. Specifically, 300 randomly generated weight vectors are utilized for the coils, minimizing the risk of converging to local minima.

\section{Data and Experiments}

\subsection{Data}
Electromagnetic simulations are conducted using a commercial finite element method (FEM)-based solver (Ansys HFSS, Canonsburg, PA, USA). The transmit array consisted of eight loop elements arranged in a single-row configuration, mounted on a 28 cm diameter cylinder to accommodate a receive coil. Each element measured 16×10 cm² and is spaced approximately 1 cm apart. A standard human body model from Ansys is employed, scaled to represent 10 scenarios reflecting average male and female dimensions across five countries. The transmit array operated at 298 MHz, corresponding to the 7T Larmor frequency. Each model maintained consistent voxel size, and the simulations produced 64 volumes of 8-channel $B_{1}^{+}$ magnetic fields with different scale sizes, each with dimensions 101×101×71×8. Additionally, the mass density data for the 64 head models is included.

To prepare the $B_{1}^{+}$ fields for analysis, several pre-processing steps are applied. For each head model, we reviewed the 71 slices and selected 32 valid cases to exclude simulation errors. Binary masks are then generated based on the mass density map of each subject, preserving regions of interest (e.g., skull and brain) while excluding surrounding air. Data augmentation techniques, such as rotation, are applied to increase the size of the data set, eventually producing 24576 masked $B_{1}^{+}$ slices for further use. The $B_{1}^{+}$ fields for each channel are configured in quadrature mode. During each experimental fold, the weights for training are derived using adaptive moment estimation to minimize the loss defined in Eq.~\ref{b}. Random initial weights are used to ensure diverse starting points for optimization, and the best performing weights are selected as training close-to-truth upper bounds. 

\subsection{Experiments Design}
For comparison, a single CPU-based MLS optimization and uCNN~\cite{11Kilic2024} using the same $B_{1}^{+}$ map inputs are performed for each sample in every fold. The weights obtained from MLS serve as a benchmark for numerical evaluation, providing a reference against which to assess the proposed approach. The uCNN serves as a comparable deep learning structure to fight against our ResNet backbone. Runtime is also recorded to facilitate a direct comparison of computational efficiency between methods.

For our proposed method, ResNet18~\cite{13He2016} is used to predict RF coil weights from $B_{1}^{+}$ maps by learning the mapping between the input data and the upper bound weights. The architecture uses an input size of 101×101×32 and produces an output of 32 weights. It starts with an initial convolutional layer, followed by four stages of residual blocks.

Each stage consists of two Basic-blocks, where each block includes two 3×3 convolutional layers, batch normalization, and ReLU activation. The feature map sizes are 64, 128, 256, and 512, doubling at each stage. Downsampling is achieved through a two-step pattern in the first block of each stage, using 1×1 convolutions to match dimensions. The final layers involve adaptive average pooling to a 1×1 output, followed by a fully connected layer that maps to 32 output classes, ensuring effective gradient propagation and robust training.

During training, the loss function is defined as the mean square error between the predicted and reference RMSE:

\begin{equation}\label{RMSE}
\begin{split}
\text{RMSE} = \sqrt{ \frac{ \left\| \left| \mathbf{A} \mathbf{b} \right| - \mathbf{m} \right\|_{\mathbf{w}}^2 }{ N_{\text{voxel}} } }
\end{split}
\end{equation}

\begin{equation}\label{loss}
\begin{split}
\text{loss} = \frac{1}{N_{\text{slice}}} \sum_{i=1}^{N_{\text{slice}}} \left\{ \left| {\text{RMSE}_{\text{pred}}}^{(i)} - {\text{RMSE}_{\text{ref}}}^{(i)} \right| \right\}
\end{split}
\end{equation}

\noindent where RMSE is calculated from Eq.~\ref{b}, $N_{\text{voxel}}$ is the number of voxels in the current slice, ${\text{RMSE}_{\text{pred}}}^{(i)}$ is the RMSE calculated from predicted weights on the $i_{th}$ slice, ${\text{RMSE}_{\text{ref}}}^{(i)}$ is the RMSE calculated from reference weights and $N_{\text{slice}}$ is the number of slices.

The entire dataset is randomly divided into training, validation, and testing subsets in an 8:1:1 ratio, with a batch size of 16. To ensure robustness and minimize the impact of random variations, the dataset separation and experimental process (training and testing) are repeated five times. Adaptive Moment Estimation (Adam)~\cite{17Kingma2014} is used as the optimizer, with an initial learning rate of \( 10^{-3} \) that decayed by \( 50\% \) every 50 epochs, over a total of 200 epochs. All training and testing are performed using PyTorch on a NVIDIA GeForce RTX A6000 GPU with CUDA 12.3. 

\subsection{Post-processing with NFD (Optional)}
To overcome the limitations of relying solely on RMSE for optimization, we introduce NFD, an optional post-processing step inspired by DCGANS~\cite{Goodfellow2014, Radford2015} to detect and penalize non-uniform artifacts in the output fields. These artifacts, representing undesirable anomalies in the output, can persist even when RMSE values appear acceptable, ultimately compromising the quality of the final results.

Uniform and non-uniform cases are derived using the MLS method for NFD training. To generate these cases, we first computed the RF shimming weights using MLS optimization. For uniform cases, the $|B_{1}^{+}|$ maps from all eight channels are visually verified to confirm the absence of artifacts. For non-uniform cases, we manually selected instances where $|B_{1}^{+}|$ maps exhibited significant artifacts or voids, indicative of non-uniform anomalies. After this selection process, we finalized a dataset containing 384×64 uniform slices and 384×64 non-uniform slices, ensuring a balanced and representative dataset for effective NFD training. During training, the NFD learns to differentiate non-uniform fields from the uniform ones, identifying subtle non-uniform anomalies.

For testing the performance of NFD, a set of previously unseen non-uniform and uniform cases (3840 non-uniform cases and 4992 uniform cases), along with their corresponding ground truth labels, are provided as input to the trained NFD. The model's classification predictions are evaluated by comparing them to the true labels, allowing for an assessment of its accuracy and effectiveness in distinguishing between the two categories.

\section{Results}

\subsection{RF Shimming Prediction Results}

\begin{table}[t]
\centering
\resizebox{\textwidth}{!}{
\begin{tabular}{c|cccc|cccc}
\hline\hline
 & \multicolumn{4}{c|}{Mean RMSE [\% of Target FA]} & \multicolumn{4}{c}{Runtime(200 slices per volume)} \\
\cline{2-9}
Fold & Upper Bound & MLS$^*$ & uCNN$^\dag$ & Ours$^\dag$  & Upper Bound & MLS$^*$ & uCNN$^\dag$ & Ours$^\dag$\\
\hline
1 & 8.5674 & 9.9176 & 9.9223 & 9.0406 & $ \sim $3 (h) & 12.3927 (min) & 0.1543 (s) & 0.1391 (s) \\
2 & 8.5640 & 9.8579 & 9.9303 & 9.0365 & $ \sim $3 (h) & 13.7560 (min) & 0.1544 (s) & 0.1386 (s) \\
3 & 8.5201 & 9.8505 & 9.8965 & 8.9969 & $ \sim $3 (h) & 13.6140 (min) & 0.1535 (s) & 0.1380 (s) \\
4 & 8.5052 & 9.7818 & 9.0496 & 9.0075 & $ \sim $3 (h) & 15.0277 (min) & 0.1529 (s) & 0.1393 (s) \\
5 & 8.5455 & 9.8719 & 9.0404 & 9.0071 & $ \sim $3 (h) & 11.2699 (min) & 0.1538 (s) & 0.1385 (s) \\
\hline\hline
\end{tabular}
}
\caption{Comparison of metrics among the CPU-based MLS method, our method, uCNN and upper bounds gained from Adam. In each fold, both methods are evaluated on the same dataset with identical splits. The Mean RMSEs are computed and expressed as [\% of Target FA], where a lower RMSE indicates better homogeneity. Runtime is measured for the same tasks, and the average runtime for 200 slices is calculated based on the MPRAGE sequence.\\
$^*$Calculated on CPU.\quad
$^\dag$Trained and tested on GPU.}
\label{tab:table}
\end{table}

\begin{figure*}[t]
\begin{center}
\includegraphics[width=1\linewidth]{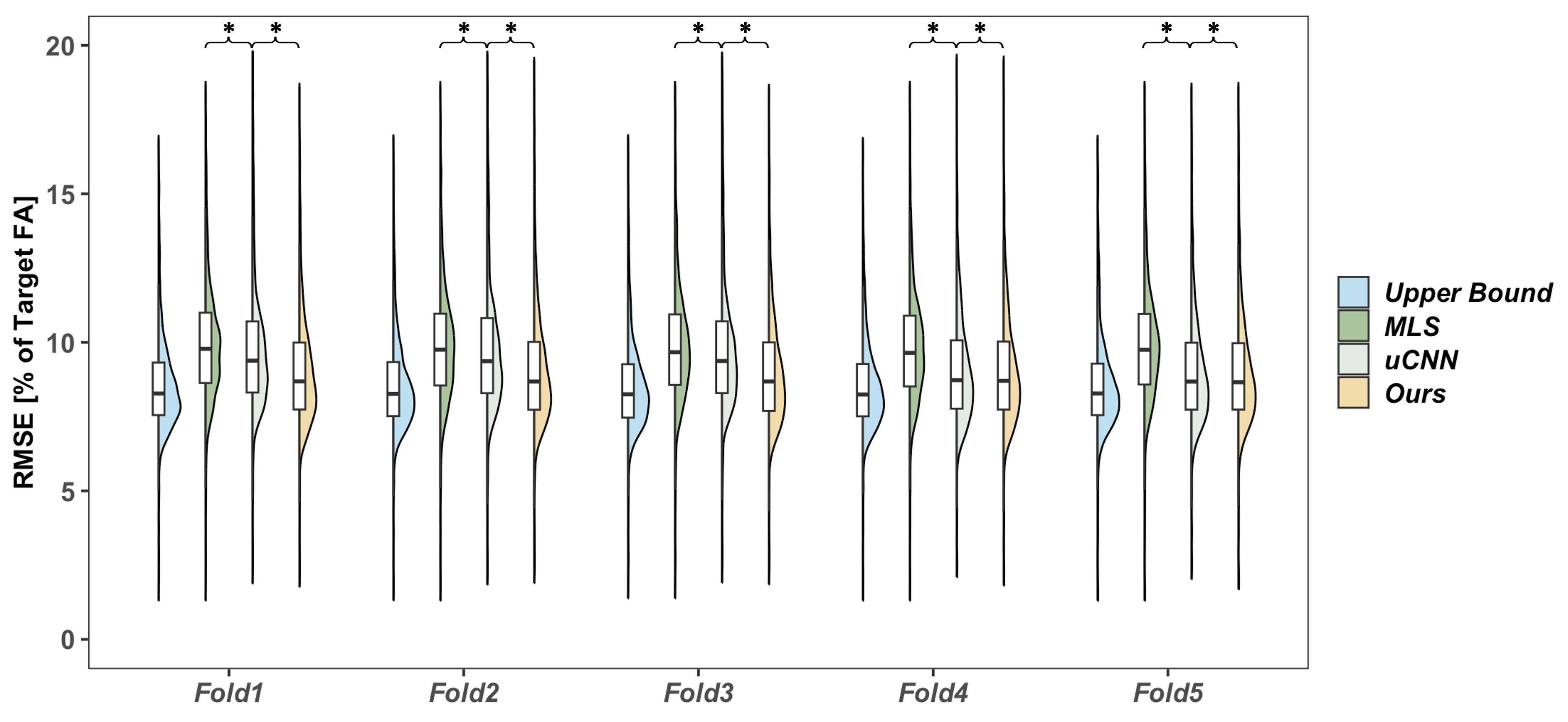}
\end{center}
\caption{The RMSE results [\% of target FA] of the upper bound, MLS method, our method and uCNN are illustrated using box plots and violin plots. Five folds of testing results are drawn independently in five groups in the figure. Illustrated by $*$, a significant difference with $p < 0.05$ is ensured.}
\label{fig:Figure_3_clean}
\end{figure*}

The performance of the proposed deep learning-based optimization strategy is compared with the traditional MLS method and uCNN across five folds using different dataset splits. The evaluation focused on Mean RMSE, runtime per volume, and their respective distributions, as shown in Tab.~\ref{tab:table} and Fig.~\ref{fig:Figure_3_clean}. We defined 200 slices as a test volume for runtime calculation, based on the Magnetization-Prepared Rapid Gradient-Echo (MPRAGE) sequence~\cite{Mugler1990}, which recommends this configuration for comprehensive and high-resolution imaging.

\subsection{Post-processing Evaluation}
\begin{figure*}[h]
\centering
\includegraphics[width=1\linewidth]{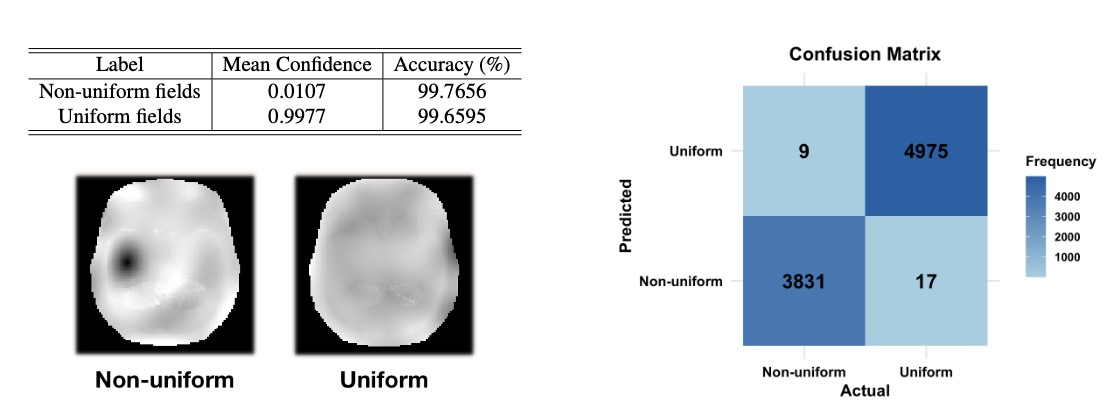}
\caption{(Left) The top section illustrates the classification performance of the NFD for non-uniform and uniform cases, while the bottom displays examples of the corresponding output fields. The table shows the mean confidence scores and accuracy percentages for each group. (Right) The confusion matrix visualizes the frequency of predictions for non-uniform and uniform cases.}
\label{fig:Figure_4_clean}
\end{figure*}

The performance of the optional post-processing step in classifying non-uniform and uniform cases is summarized in Fig.~\ref{fig:Figure_4_clean}. The evaluation is conducted using a test set of previously unseen fields, and the classification results are presented as a table and confusion matrix.

\section{Discussion}
\subsection{RF Shimming Prediction Analysis}
The proposed deep learning-based optimization strategy demonstrates substantial advantages over the traditional MLS method and uCNN in both accuracy and runtime. As shown in Tab.~\ref{tab:table}, our method consistently achieves lower Mean RMSE values across all five folds, ranging from 9.0365 to 9.0406 [\% of Target FA], compared to the MLS method's range of 9.8505 to 9.8719 [\% of Target FA] and uCNN's range of 9.0404 to 9.9303 [\% of Target FA]. This reduction in RMSE indicates improved magnetic field homogeneity, which is crucial for high-quality imaging in UHF MRI.


The runtime analysis further underscores the efficiency of the proposed approach. The runtime of our method and uCNN are close since they are both using trained deep learning models to predict the weights. However, the CPU-based MLS method requires an average of 12 to 15 minutes per volume (200 slices), whereas our method, trained on GPU with a total training time of about one hour, processes the same volume in only about 0.14 seconds. Notably, the speed advantage of our method will become even more pronounced as the number of slices or subjects increases, since the deep learning model requires only a one-time training phase and can then be applied rapidly to any new data. This dramatic improvement in computational speed makes our approach well-suited for real-time and large-scale applications, effectively addressing a major limitation of traditional optimization techniques. Additionally, the narrow interquartile ranges observed in the RMSE distributions (Fig.~\ref{fig:Figure_3_clean}) indicate that our method is not only more accurate, but also more robust across different folds and dataset splits.

\subsection{NFD Post-processing Analysis}
The post-processing step, incorporating the NFD, adds a crucial layer to address non-uniform artifacts that may persist despite acceptable RMSE values. These artifacts, which compromise the quality of the output fields, require explicit detection and penalization to ensure clinically useful results.

The classification performance of the NFD is summarized in Fig.~\ref{fig:Figure_4_clean}. It achieves a high classification accuracy of 99.77\% for non-uniform cases and 99.66\% for uniform cases, with mean confidence scores of 0.0107 and 0.9977, respectively. This demonstrates the NFD’s robustness in differentiating between non-uniform and uniform outputs, ensuring its reliability in identifying and penalizing undesirable anomalies.

The inclusion of the NFD would enhance the overall optimization process. While RMSE provides a quantitative measure of magnetic field homogeneity, it alone cannot guarantee the absence of non-uniform artifacts. The NFD addresses this limitation by offering an additional evaluation metric and refining the optimization process to ensure that final outputs meet both quantitative and qualitative standards. This optional step bolsters the framework's ability to produce artifact-free results suitable for clinical use and further improves the reliability of RF shimming weights in real-world applications.

\subsection{Limitation of Simulation-Only Training}
A key limitation of the present study is its exclusive reliance on EM simulations for both the training and evaluation of the proposed deep learning framework. Although simulated $B_1^+$ fields provide a controlled and reproducible environment for magnetic and electric field generation, simulation-based validation cannot fully capture the complexity and variability encountered in real clinical scenarios, such as physiological motion, coil loading differences, hardware imperfections, and the presence of unexpected artifacts. As a result, models trained solely on simulated data may exhibit limited generalizability when applied to real patient data, potentially affecting clinical applicability~\cite{rv_realdata1, rv_realdata2}. Therefore, a potential future plan would be evaluating the performance on real human subjects. For example, we could test the current model on measured $B_1^+$ maps and further improve robustness by incorporating real data into training.

\subsection{Scalability and Generalization}
While the proposed framework demonstrates strong performance at 7T, its application to other field strengths (including 3T, 9.4T, and 10.5T) should still be considered. UHF MRI systems like 7T face significant \(B_{1}^{+}\) inhomogeneities due to shorter RF wavelengths, making advanced shimming essential. In contrast, 3T systems encounter fewer inhomogeneities but still require optimization, especially for abdominal and pelvic MRI.

Adapting the framework to other static fields would require modifications to the input data, trained on \(B_{1}^{+}\) fields measured at 7T, and recalibration of the NFD at the post-processing stage to detect subtle inhomogeneities typical of lower field strengths. While non-uniform artifacts are less common at 3T, the framework could target gradient-induced inhomogeneities instead. The high computational efficiency of our method, which processes 200 slices in just 0.139 seconds, makes it particularly suitable for 3T systems, where speed is critical for high-throughput clinical workflows. 

Our future work should validate the framework on datasets from 3T, 9.4T, and 10.5T systems, exploring its generalizability across different field strengths. Incorporating specific absorption rate (SAR) limits and handling dynamic scenarios, such as patient motion, would further enhance its versatility for both research and clinical applications.

\section{Conclusions}
The proposed Fast-RF-shimming framework offers a transformative solution to RF shimming in UHF MRI, addressing challenges in both accuracy and efficiency. By integrating adaptive optimization, a ResNet-based architecture and an option post-processing step, the method outperforms traditional MLS optimization, achieving lower RMSE values, faster runtimes, and effective artifact mitigation. These advancements demonstrate its potential for enhancing real-time shimming workflows and improving diagnostic imaging quality in ultra-high-field MRI.

Future directions include adapting the framework to different magnetic field strengths, such as 3T, 9.4T, and 10.5T, and validating its scalability across diverse MRI systems. Expanding its capabilities to account for dynamic scenarios, such as patient motion, and incorporating specific absorption rate (SAR) constraints will further enhance its applicability. With its demonstrated efficiency, robustness, and adaptability, our method represents a significant step forward in optimizing MRI performance for both clinical and research applications.

\section{Acknowledgments}
This research was supported by NIH R01DK135597 (Huo), DoD HT9425-23-1-0003 (HCY), R01 EB031078 (Yan), R21 EB029639 (Yan), R03 EB034366 (Yan), S10 OD030389, and NIH NIDDK DK56942 (ABF). This work was also supported by Vanderbilt Seed Success Grant, Vanderbilt Discovery Grant, and VISE Seed Grant. This project was supported by The Leona M. and Harry B. Helmsley Charitable Trust grant G-1903-03793 and G-2103-05128. This research was also supported by NIH grants R01EB033385, R01DK132338, REB017230, R01MH125931 and NSF 2040462. We extend gratitude to NVIDIA for their support by means of the NVIDIA hardware grant. This works was also supported by NSF NAIRR Pilot Award NAIRR240055. This manuscript has been co-authored by ORNL, operated by UT-Battelle, LLC under Contract No. DE-AC05-00OR22725 with the U.S.Department of Energy.

This paper describes objective technical results and analysis. Any subjective views or opinions that might be expressed in the paper do not necessarily represent the views of the U.S. Department of Energy or the United States Government.  

\section{Declaration of generative AI and AI-assisted technologies in the writing process}

During the preparation of this work, the author(s) used ChatGPT4/ChatGPT4o in order to check for writing errors and perform appropriate editing and refinement. After using this tool, the author(s) reviewed and edited the content as needed and take(s) full responsibility for the content of the published article.

\bibliographystyle{elsarticle-num_clean} 
\bibliography{main_clean}

\newpage
\listoffigures

\end{document}